# NON-RIGID IMAGE REGISTRATION USING SELF-SUPERVISED FULLY CONVOLUTIONAL NETWORKS WITHOUT TRAINING DATA

*Hongming Li, Yong Fan*

Department of Radiology, Perelman School of Medicine, University of Pennsylvania, Philadelphia, PA, 19104, USA

## ABSTRACT

A novel non-rigid image registration algorithm is built upon fully convolutional networks (FCNs) to optimize and learn spatial transformations between pairs of images to be registered in a self-supervised learning framework. Different from most existing deep learning based image registration methods that learn spatial transformations from training data with known corresponding spatial transformations, our method directly estimates spatial transformations between pairs of images by maximizing an image-wise similarity metric between fixed and deformed moving images, similar to conventional image registration algorithms. The image registration is implemented in a multi-resolution image registration framework to jointly optimize and learn spatial transformations and FCNs at different spatial resolutions with deep self-supervision through typical feedforward and backpropagation computation. The proposed method has been evaluated for registering 3D structural brain magnetic resonance (MR) images and obtained better performance than state-of-the-art image registration algorithms.

## 1. INTRODUCTION

Medical image registration is typically formulated as an optimization problem to seek a spatial transformation that establishes pixel/voxel correspondence between a pair of fixed and moving images. Recently, deep learning techniques have been used to build prediction models of spatial transformations for image registration under a supervised learning framework [1-3], besides learning image features for image registration using stacked autoencoders [4]. These methods are designed to predict spatial relationship between image pixel/voxels from a pair of images based on their image patches. The learned prediction model can then be applied to images pixel/voxel-wisely to achieve an overall image registration.

The prediction based image registration algorithms typically adopt convolutional neural networks (CNNs) to learn informative image features and a mapping between the learned image features and spatial transformations that register images in a training dataset [1-3]. Similar to most machine learning tasks, the quality of training data plays an important role in the prediction based image registration. A variety of strategies have been proposed to build training data, specifically the spatial transformations [1-3]. However, a prediction based image registration model built upon such training datasets is limited to estimating spatial transformations captured by the train datasets themselves.

Inspired by spatial transformer network (STN) [5], deep CNNs in conjunction with STNs have been proposed recently to learn prediction models for image registration in an unsupervised fashion [6, 7]. In particular, DirNet learns CNNs by optimizing an image similarity metric between fixed and transformed moving images to estimate 2D control points of cubic B-splines for representing spatial transformations [6]; ssEMnet estimates coarse-grained deformation fields at a low spatial resolution and uses bilinear interpolation to obtain dense spatial transformations for registering 2D images by optimizing an image similarity metric between feature maps of the fixed and transformed moving images [7]. However, the coarse-grained spatial transformation measures may fail to characterize fine-grained deformations between images.

Building upon fully convolutional networks (FCNs) that facilitate voxel-to-voxel learning [8], we propose a novel deep learning based non-rigid image registration framework to learn spatial transformations between pairs of images to be registered [9]. Different from most learning based registration methods that reply on training data, our method directly trains FCNs to estimate voxel-to-voxel spatial transformations for registering images by maximizing their image-wise similarity metric, similar to conventional image registration algorithms. To account for potential large deformations between images, a multi-resolution strategy is adopted to jointly learn spatial transformations at different spatial resolutions. The image similarity measures between the fixed and deformed moving images are evaluated at different image resolutions to serve as deep self-supervision. Since our method simultaneously optimizes and learns spatial transformations for the image registration in an unsupervised fashion, the registration of pairs of images also serves as a training procedure, and the trained FCNs can be directly adopted to register new images using feedforward computation. The proposed method has been evaluated based on 3D structural brain MR images.

## 2. METHODS

### 2.1. Image registration driven by image similarity metric

Given a pair of fixed image $I_f$ and moving image $I_m$, the task of image registration is to seek a spatial transformation that establishes pixel/voxel-wise spatial correspondence between the two images. Since the spatial correspondence can be gauged with a surrogate measure, such as an image intensity similarity, the image registration task can be formulated as an optimization problem to identify a spatial transformation that maximizes the image similarity measure between the fixed image and transformed moving image. For non-rigid image registration, the spatial transformation is often characterized by a dense deformation field $D_v$ that encodes displacement vectors between spatial coordinates of $I_f$ and their counterparts of $I_m$.

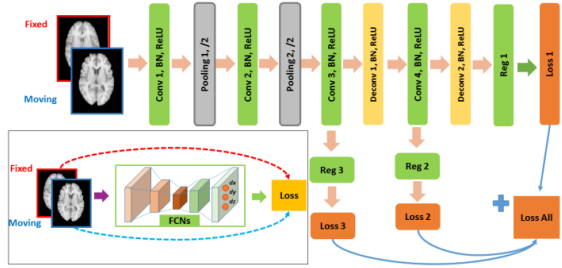

Fig. 1. Overall architecture of the image registration framework and an example FCNs for voxel-to-voxel regression of deformation fields in a multi-resolution image registration framework.

Regularization techniques are usually adopted in image registration algorithms to obtain spatially smooth and physically plausible spatial transformations. In this study, we adopt a total variation based regularizer [10]

$$R(D_v) = \sum_{n=1}^{N} \|\nabla D_v(n)\|_1, \quad (1)$$

where $N$ is the number of pixel/voxels in the deformation field. In general, the image registration problem is formulated as

$$\min_{D_v} -S\left(I_f(v), I_m(D_v \circ v)\right) + \lambda R(D_v), \quad (2)$$

where $v$ represents spatial coordinates of pixel/voxels in $I_f$, $D_v \circ v$ represents deformed spatial coordinates of pixel/voxels by $D_v$ in $I_m$, $S(I_1, I_2)$ is an image similarity measure, $R(D)$ is a regularizer on the deformation field, and $\lambda$ controls the trade-off between the image similarity measure and the regularization term.

**2.2. Image registration using FCNs**

To solve the image registration optimization problem, we build a deep learning model using FCNs to learn informative image feature representations and a mapping between the feature representations and the spatial transformation between images at the same time. The registration framework of our method is illustrated in Fig.1 (bottom left). In particular, each pair of fixed and moving images are concatenated as an input image with two channels to the deep learning model for learning spatial transformations that optimize image similarity measures between the fixed and transformed moving images. The deep learning model comprises FCNs with de/convolutional (Conv) layers, batch normalization (BN) layers, activation (ReLU) layers, pooling layers, and multi-output regression layers. Particularly, each of the regression layer (Reg) is implemented as a convolutional layer whose output has the same size of the input images in the spatial domain and multiple channels for encoding displacements in different spatial dimensions of the input images.

The pooling operation is usually adopted in CNNs to obtain translation-invariant features and increase reception fields of the CNNs, as well as to decrease the spatial size of the CNNs to reduce the computational cost. However, the multi-output regression layers after pooling operations produce coarse outputs which have to be interpolated to generate deformation fields at the same spatial resolution of the input images [6, 7]. An alternative way to obtain fine-grained deformation fields is to stack multiple convolutional layers without any pooling layers. However, such a network architecture would have more parameters to be learned and decrease the efficiency of the whole network. In this study, we adopt deconvolutional operators for upsampling [8], instead of choosing a specific interpolation scheme, such as cubic spline and bilinear interpolation [6, 7]. Our network architecture also naturally leads to a multi-resolution image registration that has been widely adopted in conventional image registration algorithms.

In this study, normalized cross-correlation (NCC) is used as the image similarity metric between images, and the total variation based regularizer as formulated by Eq. (1) is adopted to regularize the deformation fields. Therefore, the loss layer evaluates the registration loss between the fixed and deformed moving images as formulated by Eq. (2).

**2.3. Multi-resolution image registration with deep self-supervision**

Our multi-resolution image registration method is built upon FCNs with deep self-supervision, as illustrated in Fig. 1 (top right). Particularly, the first 2 pooling layers in conjunction with their preceding convolutional layers progressively reduce the spatial size of the convolutional networks so that informative image features can be learned by the 3[rd] convolutional layer to predict voxel-wise displacement at the same spatial resolution of downsampled input images. And the subsequent deconvolutional layers learn informative image features for predicting spatial transformations at higher spatial resolutions.

Similar to conventional multi-resolution image registration algorithms, the similarity of registered images at different resolutions is maximized in our network to serve as deep supervision [11], but without the need of supervised deformation field information. Such a supervised learning with surrogate supervised information is referred to as self-supervision in our study.

Different from conventional multi-resolution image registration algorithms in which deformation fields at lower-resolutions are typically used as initialization inputs to image registration at a higher spatial resolution, our deep learning based method jointly optimizes deformation fields at all spatial resolutions with a typical feedforward and backpropagation based deep learning setting. As the optimization of the loss function proceeds, the parameters within the network will be updated through the feedforward computation and backpropagation procedure, leading to improved prediction of deformation fields. It is worth noting that *no training deformation field information is needed for the optimization*, and self-supervision through maximizing image similarity with smoothness regularization of deformation fields is the only force to drive the optimization. The trained network can be directly used to register a pair of images, and any of them can be the fixed image.

### 2.4. Network training

Given a set of $n$ images, we could obtain $n(n-1)$ pairs of fixed and moving images, such that every image can serve as a fixed image. Pairs of images are registered using following parameters. As illustrated in Fig. 1, 32, 64, 128, and 64 kernels are used for Conv layer 1, 2, 3, and 4 respectively, with kernel size 3 and stride 2. For pooling layers, kernel size is set to 3, and stride 2. 64 and 32 kernels are used for Deconv layer 1 and 2 respectively, with kernel size 3 and stride 2. Three kernels are used in the regression layers 1, 2 and 3 to obtain 3D deformation fields. The total loss is calculated as a weighted sum of loss of the 3 loss layers, with weight coefficients 1, 0.6, and 0.3 assigned to the loss layers 1, 2, and 3 respectively.

We implement alternative network architecture without pooling layers for performance comparison. Particularly, Conv layers 1 to 3, one regression layer, and one loss layer are kept. The Conv layers have the same parameters as described above. Moreover, alternative network architecture with pooling layers and additional one interpolation layer is also implemented as an image registration model with coarse-grained spatial transformation, and tri-linear interpolation is adopted to upsample the coarse-grained deformation fields to the original spatial resolution.

The registration models are built using Tensorflow [12]. Adam optimization technique is adopted to train the networks, with learning rate set to 0.001. The networks are trained on one Nvidia Titan Xp GPU with 10000 iteration steps. The trained FCNs could be directly used to register new images with feedforward computation.

### 3. RESULTS

The 1st dataset used in this study was obtained from the Alzheimer's Disease Neuroimaging Initiative (ADNI) database (http://adni.loni.usc.edu). Particularly, baseline MRI data of 959 subjects were obtained from the ADNI Go & 2, and 817 subjects from the ADNI 1. T1-weighted MRI scans of all the subjects were registered to the Montreal Neurological Institute (MNI) space using affine registration, and then a 3D bounding box of size 32×48×48 was adopted to extract hippocampus regions for each subject, as similarly did in a hippocampus segmentation study [13]. In addition, 100 T1 images with hippocampus segmentation labels were obtained from a preliminary release of the EADC-ADNI harmonized segmentation protocol project [14]. These images with hippocampus labels were used to evaluate image registration performance based on an overlapping metric between the hippocampus labels of registered images.

The 2nd dataset used was LPBA40 in delineation space [15]. The LPBA40 consists of 40 T1-weighted MRI brain images and their label images, each with 56 brain regions. All of these MRI brain images were registered to MNI152 space at a spatial resolution of 2x2x2 mm$^3$ after the image intensity was normalized using histogram matching, and their label images were transformed to MNI152 space accordingly. These MRI images with their labels (excluding cerebellum and brainstem) were used to evaluate image registration performance based on an overlapping metric between the labels of registered images.

We compared our method with the deep learning methods for learning coarse-grained deformation [6, 7] and ANTs [16] based on the same datasets.

For the ADNI dataset, all the deep learning based image registration models were trained based on the ADNI GO & 2 dataset, and evaluated based on the ADNI 1 dataset. The model was trained with a batch size of 64 (32 for the network without pooling layer due to GPU memory limit). In the testing phrase, 9600 pairs of images were randomly selected to test the performance of different models.

The image registration performance obtained by different deep learning models on the ADNI dataset was evaluated using the NCC measure. The mean NCC measures obtained by the model with and without pooling layers, and the deep model with multi-resolution strategy were 0.616, 0.848, 0.905, respectively, indicating that the coarse-grained deformation fields obtained could not capture fine-grained image correspondence. Our model obtained the best performance, demonstrating that the multi-resolution model could extract more informative image representations and better learn spatial relationship between images.

To compare our method with ANTs, we randomly selected one image from the EADC-ADNI dataset as the fixed image, and registered all other images to it. For our method, we directly used the trained deep learning model to register these images, and ANTs registered the images with following command: ANTS 3 -m CC [fixed, moving, 1, 2] -o output -t SyN [0.25] -r Gauss [3,0] -i 100x100x10 --number-of-affine-iterations 100x100x100 --use-Histogram-Matching 1. The deformation fields obtained were applied to register their corresponding hippocampus labels. The overlapping between the fixed and registered moving labels was measured using Dice index.

The Dice index values of the hippocampus labels between images before and after registration by ANTs and our method were $0.654 \pm 0.062$, $0.762 \pm 0.057$, and $0.798 \pm 0.033$, respectively. These results indicate that the proposed method could identify better spatial correspondence between images. Moreover, it took ~ 1 minute to register two images by ANTs on one CPU (AMD Opteron 4184 @ 2.80Ghz), while only ~50 ms by our model on one Titan Xp GPU.

We further compared our method and ANTs using the LPBA40 dataset. In particular, our deep learning based image registration model was trained based on 30 images, and the remaining 10 images were used as the testing images. In the training phrase, image pairs were randomly selected from 30 training images, and the batch size was set to 8. In the testing phrase, each testing image was used as the fixed image, and all the other 9 testing images were registered to it using the trained deep learning model. The ANTs algorithm was applied directly to register the testing images in the same manner, with the same ANTs command as described above. The overlap between deformed label and ground truth label for 54 regions of the testing images were calculated to evaluate the registration performance.

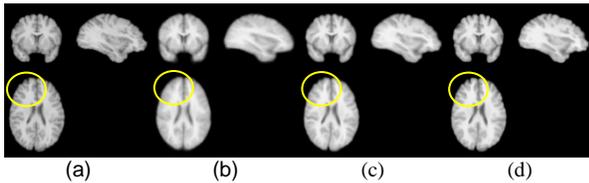

(a) (b) (c) (d)

Fig. 2. Mean brain image before and after registration by different methods. (a) Fixed image, (b) mean of images before registration, (c) mean of registered images by ANTs, (d) mean of registered image by the proposed method.

The mean images before and after registration by different methods with one randomly selected testing image as the fixed image are shown in Fig. 2. As shown in Fig. 2b, the mean of images before registration looks blurry, and the means of registered images in Fig. 2c and 2d maintain detailed image textures, and the one obtained by our method has sharper contrast than that obtained by ANTs visually.

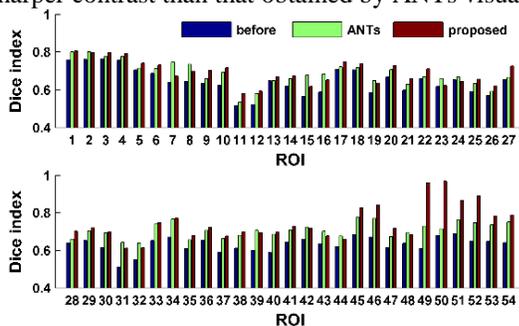

Fig. 3. Dice index for 54 ROIs between all testing image pairs from the LPBA40 dataset before and after registration using ANTs and the proposed method.

The mean Dice index values of all the testing images for the 54 regions of interest (ROIs) are illustrated in Fig. 3. The Dice index values after registration were significantly higher than that before registration. For 35 out of 54 ROIs, their Dice index values obtained by our method were significantly higher than those obtained by the ANTs (Wilcoxon signed rank test, $p<0.02$). The mean Dice index across all the ROIs of the testing images before and after registration by the ANTs and our method were 0.639, 0.697, and 0.720 respectively. No optimization was performed by our method for registering the testing images, and it took ~200 ms to register a pair of images.

## 4. CONCLUSION

We present a novel deep learning based non-rigid image registration algorithm to learn spatial transformations between pairs of images to be registered. The experimental results based on 3D structural MR images have demonstrated that our method could obtain promising image registration performance with respect to both image registration accuracy and computational speed.

### ACKNOWLEDGEMENTS

This work was supported in part by National Institutes of Health grants [EB022573, CA223358, MH107703, DK114786, DA039215, and DA039002].